\documentclass[lettersize, journal]{IEEEtran}
\IEEEoverridecommandlockouts
\usepackage{makecell}
\usepackage{multirow}
\usepackage{booktabs}
\usepackage{threeparttable}

\usepackage{colortbl}
\usepackage{hhline}
\usepackage{color}
\usepackage[dvipsnames, svgnames, x11names]{xcolor}
\usepackage{amsmath,amsfonts}
\usepackage{soul}

\usepackage{algorithm}
\usepackage[english]{babel}
\usepackage{vcell}
\usepackage{tablefootnote}
\usepackage[utf8]{inputenc}

\usepackage{caption}

\usepackage{algorithmic}

\makeatletter
\newcommand{\removelatexerror}{\let\@latex@error\@gobble}
\makeatother

\usepackage{array}

\usepackage[caption=false,font=normalsize,labelfont=sf,textfont=sf]{subfig}
\usepackage{textcomp}
\usepackage{stfloats}
\usepackage{verbatim}
\usepackage{graphicx}
\usepackage{epsfig}
\usepackage{hyperref}
\DeclareUnicodeCharacter{2061}{}
\usepackage[backend=bibtex,style=ieee,sorting=none,isbn=false,url=false,doi=false]{biblatex}
\title{Theoretical Model Construction of Deformation-Force for Soft Grippers Part I: Co-rotational Modeling and Force Control for Design Optimization}

\author{Huixu Dong, Haotian Guo, Sihao Yang, Chen Qiu, Jiansheng Dai,~\IEEEmembership{Fellow,~IEEE}, I-Ming Chen, ~\IEEEmembership{Fellow,~IEEE}
\thanks{\noindent Huixu Dong, Haotian Guo, Sihao Yang are with Robot Perception
and Grasp Laboratory(Grasp Lab), Zhejiang University, Hangzhou 310058,
China (e-mail: huixudong@zju.edu.cn). I-Ming Chen is with Robotics Research Center, Nanyang Technological University, Singapore 639798. Chen Qiu is with Maider Medical Industry Equipment Co., Itd, China 317607. Jiansheng Dai is with Shenzhen Key Laboratory of Biomimetic Robotics and Intelligent Systems, SUSTech Institute of Robotics, Southern University of Science and Technology, Shenzhen, 518055, China, and Centre for Robotics Research, Department of Engineering, King's College London Strand, London WC2R 2LS, UK.}}

\vspace{-10mm}
\begin{document}
\maketitle

\begin{abstract}
Compliant grippers, owing to adaptivity and safety, have attracted considerable attention for unstructured grasping in real applications, such as industrial or logistic scenarios. However, accurately modeling the bidirectional relationship between shape deformation and contact force for such grippers, the Fin-Ray grippers as an example, remains stagnant to date. To address this research gap, this article devises, presents, and experimentally validates a universal bidirectional force-displacement mathematical model for compliant grippers based on the co-rotational concept, which endows such grippers with an intrinsic force sensing capability and offers a better insight into the design optimization. In Part \uppercase\expandafter{\romannumeral1} of the article, we introduce the fundamental theory of the co-rotational approach, where arbitrary large deformation of beam elements can be modeled. Its intrinsic principle allowing taking materials with varying stiffness, various connection types and key design parameters into consideration with few assumptions. Further, the force-displacement relationship is numerically derived, providing accurate displacement estimations of the gripper under external forces with minor computational loads. \textcolor{Purple}{The performance of the proposed method is experimentally verified through comparison with Finite Element Analysis (FEA) in simulation, obtaining a fair degree of accuracy (6\%) , and design optimization of Fin-Ray grippers is systematically investigated.} \textcolor{Blue}{Part \uppercase\expandafter{\romannumeral2} of this article demonstrating the force sensing capabilities and the effects of representative co-rotational modeling parameters on model accuracy is released in Arxiv\footnote{\href{https://arxiv.org/pdf/2303.12418.pdf}{Part \uppercase\expandafter{\romannumeral2}: https://arxiv.org/pdf/2303.12418.pdf}}.}
\end{abstract}

\textbf{\textit{Index Terms---}Compliant gripper, Optimal design, Bidirectional modeling, Grasp performance }

\section{Introduction}
\begin{figure}[tbp]
\centerline{\includegraphics[width=0.85\columnwidth]{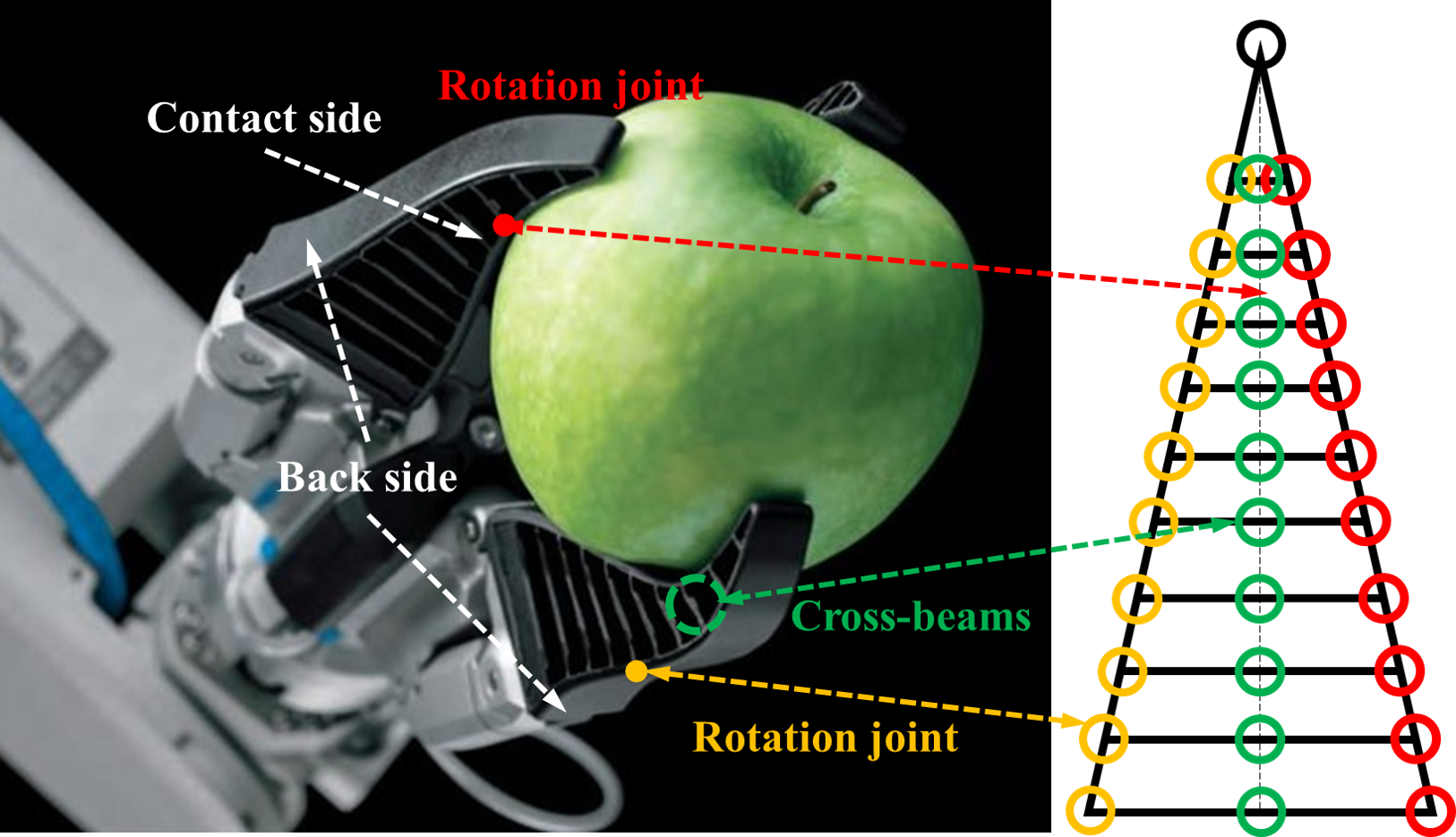}}
\caption{\small{Festo’s soft gripper based on Fin-Ray® effect and its equivalent Fin-Ray structure, \textcolor{red}{retrived from \href{https://www.festo.com/net/SupportPortal/Files/333986/Festo_MultiChoiceGripper_en.pdf}{Festo Gmbh}.} The green circles represent the crossbeams; the yellow circles indicate the joints of the back side and the red circles are the joints of the contact side.
}}
\vspace{-6mm}
\label{fig:1}
\end{figure}

\IEEEPARstart{G}{rasp} is an essential capability for most robots in practical applications \cite{ref1,ref2}. As grasping executors, compliant grippers have an obvious advantage in grasping unstructured objects, such as soft, fragile, deformable, time-varying ones \cite{ref3,ref4, GSG}. Generally, current compliant grippers can be categorized as either multiple-rigid-link grippers that perform compliant grasps or most soft grippers. To achieve compliance, rigid grippers are usually designed as under-actuated ones by combining multiple rigid links and joints with only fewer actuators \cite{ref11,ref14,ref16}. Soft grippers possess superior compliant performance in grasping arbitrary-shaped objects owing to the infinite amount of Degree of Freedom (DOF). Their deformations are usually based on materials' intrinsic mechanical properties and investigations mainly emphasize the gripper design \cite{ref18,ref26,ref27} and actuation strategies \cite{ref19}. By contrast, a Fin-Ray gripper, a migration from the rigid gripper to a soft one, achieves a good balance of compliance and payload and has attracted researchers' attention \cite{ref4,ref6}. Figure \ref{fig:1} reveals a commercialized product, Festo DAHS. 

Despite the abundant achievements in compliant grippers design, rare literature mentions their theoretical modeling due to either the non-linear models with complex parameters or the substantial amount of assumptions and constraints. \textcolor{ForestGreen}{Combining kinematic and static analyses of the under-actuated slider-crank, Yoon et al. designs a compliant rigid gripper revealing strong pinch capability with optimized parameters \cite{Force_linkage}. Despite the robust adpation towards environments, their methods is limited to linkage-based grippers. 
Shan et. al combines virtual work theory and pseudo-rigidbody model to predict the grasping force and overall grasp strength along finger surface \cite{ref4}. Similarly, 
utilizing the discrete-Cosserat-approach, Armanini et al. models the deformation of closed-chain soft robots sharing the geometrical structure as rigid counterparts and improves the Fin-Ray gripper design \cite{Force_FR}. Xu's work employs the Nerual Networks to entitle the Fin-Ray gripper with intrinsic force sensing capabilities \cite{ref6}.
However, despite the high accuracy of these models, they could either predict deformation under external forces or contact forces given the displacement of nodes. By contrast, this paper aims to theoretically construct the bi-directional mathematical model that mutually depicts the force-displacement behaviors of the compliant grippers with a fair degree of accuracy and reduced constraints, which none of the previous models dealt with.}

Inspired by the co-rotational concept \cite{ref51}, which offers an accurate compliance modeling of arbitrary large deformation of beam elements under external loads, this article first explores, presents, and verifies the constructed bidirectional force-deformation mathematical model, providing insights into design optimization of compliant grippers and force sensing capability sensor-free. \textcolor{Blue}{The Fin-Ray gripper is chosen for its generalization ability to other compliant grippers. Especially, a soft/continuum gripper can be commonly considered a Fin-Ray gripper with numerous crossbeams with low stiffness material for readily deforming. While a compliant rigid gripper with multiple joints may be regarded as a Fin-Ray gripper with several crossbeams constructed by high-stiff material. }

The foremost contribution of this article is that we uniquely model and analyze the mathematical relationships between the finger deformation and contact forces through the co-rotational theory, which can be easily generalized to other compliant grippers. We highlight the novelties in \textbf{Part I}. \textbf{First}, we devise a computational-efficient force-deformation model for the Fin-Ray grippers based on the co-rotational concept and experimentally verify its effectiveness through comparisons with FEA. \textbf{Second}, 
the influence of four key design parameters for a Fin-Ray gripper on its performance is systematically investigated for design optimization. A critical insight into the optimized trade-off of performance in specified applications has been provided. In \textbf{Part II}, the intrinsic force sensing capability based on the deformation-force model is explored.
\vspace{-1mm}

\section{Modeling and Analysis}
\subsection{Co-rotational Model}
\subsubsection{Axial Deflection Modeling}

\ 

\begin{figure}
	\centering
	\begin{minipage}[!ht]{\linewidth}
		\centering
		\includegraphics[width=0.75\columnwidth]{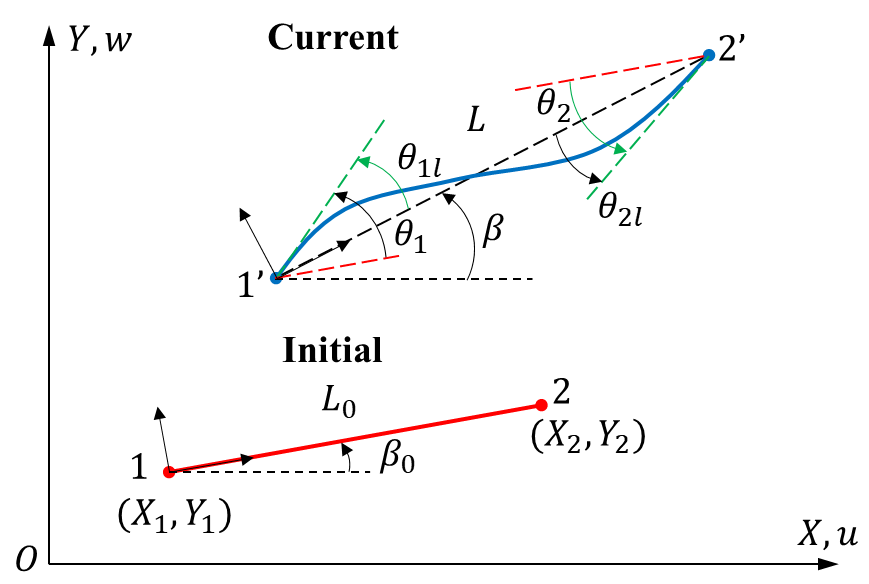}
  \vspace{-2mm}
		\caption{\small{Relationship between global coordinates and local coordinates of the nodes of each beam element.}}
		\label{fig:2}
	\end{minipage}
	\\
	\begin{minipage}[!ht]{\linewidth}
		\centering
		\includegraphics[width=0.75\columnwidth]{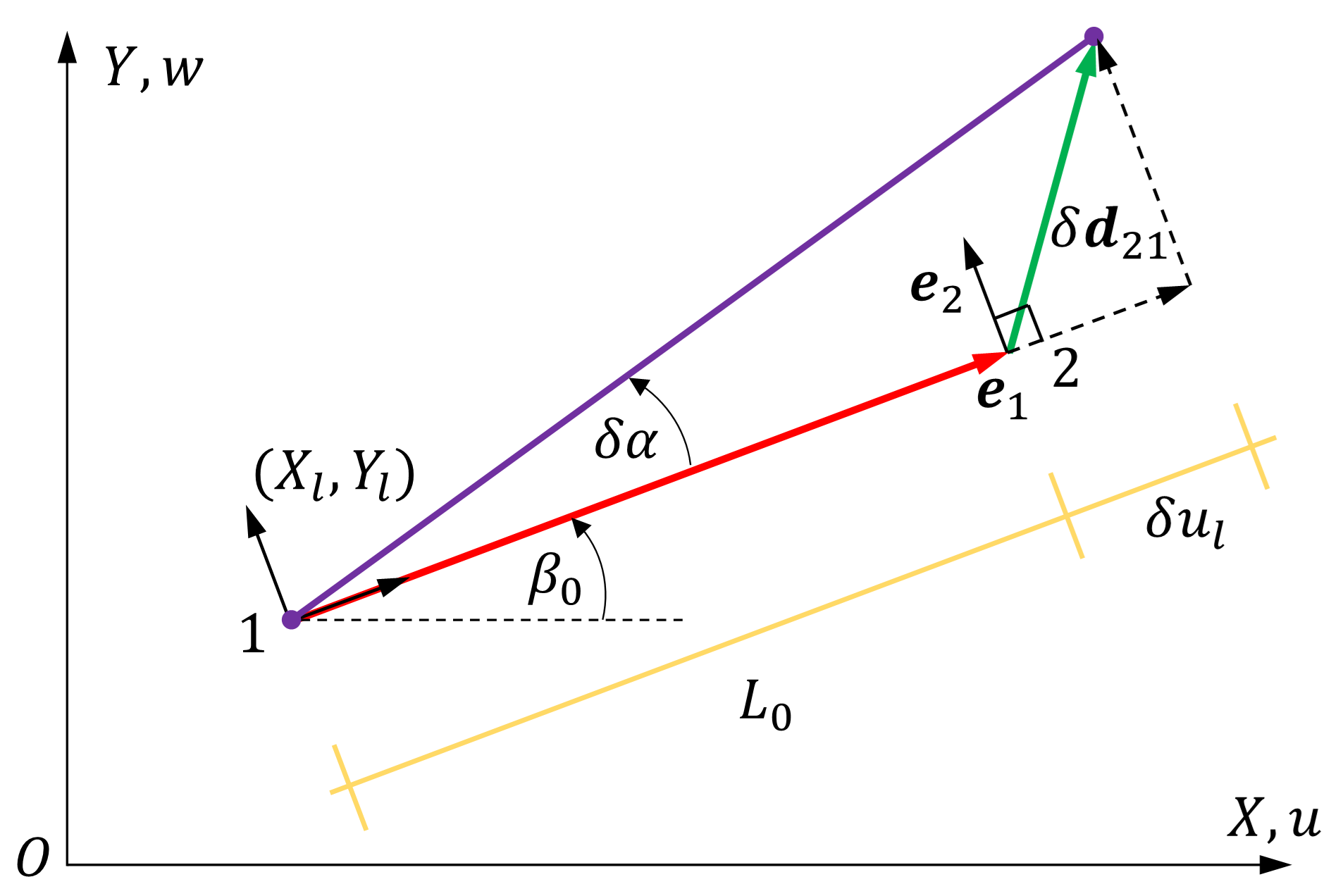}
		\caption{\small{The global displacement for a beam element. $(X_l,Y_l )$ represents the node coordinate in the local coordinate reference frame, where $(e_1,e_2)$ is the unit vector: $e_1$ is along the beam and $e_2$ is perpendicular to the beam; the purple line represents the beam after a certain displacement; $\delta d_{21}$ denotes the displacement vector; $\delta_{u_l}$ is the local translation displacement; $\delta\alpha$ represents the rotation angle of the beam.}}
		\label{fig:3}
  \vspace{-1mm}
	\end{minipage}
        \\
        \begin{minipage}[!ht]{\linewidth}
		\centering
		\includegraphics[width=0.9\columnwidth]{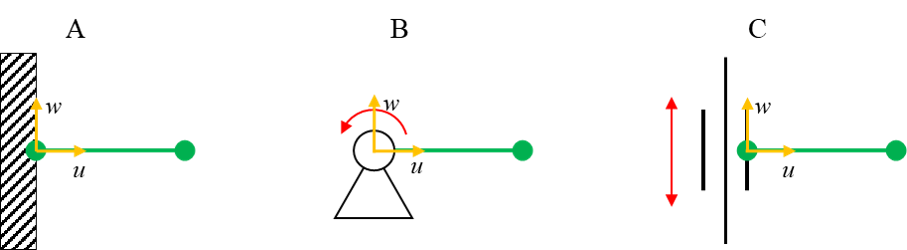}
		\caption{\small{Typical types of support nodes. The support node displacement is $(u=0,w=0,\theta=0)$ in (A), $(u=0,w=0,\theta\ne0)$ in (B), $(u=0,w\ne0,\theta=0)$ in (C).}}
		\label{fig:4}
  \vspace{-6mm}
	\end{minipage}
\end{figure}

\textcolor{Purple}{Existing research commonly assume the ribs as rigid inextensible beam and neglect the axial deformation \cite{ref4, ref6, Force_FR}.} Without losing the generality, this article provides a 2-D modeling, closely following the concept of co-rotational modeling given by Borst \cite{ref51} and Yaw \cite{ref52}, where any targeted structure can be modeled as a combination of elastic elements. For instance, a slender beam is treated as a serial combination of small beam elements. By separating the local deformation of each beam element and its rigid motion, and allowing arbitrary large motions between adjacent beam elements, the co-rotational approach can model the large deformation of objects with sufficient accuracy and efficiency. As illustrated in Figure \ref{fig:2}, each beam element has two nodes, node 1 and node 2, respectively. In the global coordinate frame $\{O, X, Y\}$, given the initial coordinates for two ends of beams ($X_1, Y_1$) and ($X_2, Y_2$), the initial angle $\beta_{0}$, and length $L_0$ of the beam can be derived accordingly. When subjected to an external load, assuming node 1 and node 2 generate the displacements, ($u_1,w_1$) and ($u_2,w_2$), respectively, the current incline angle $\beta$  and length $L$ can be derived as:
\begin{equation}
\vspace{-2mm}
L_0=\sqrt{(X_2-X_1 )^2+(Y_2-Y_1 )^2}
\label{eq:1}
\end{equation}
\begin{equation}
\begin{matrix}
\delta X=(X_2+u_2 )-(X_1+u_1 ) \\
\delta Y=(Y_2+w_2 )-(Y_1+w_1 ) \\
L=\sqrt{(\delta X)^2+(\delta Y)^2} \\
\cos ⁡\beta=\frac{\delta X}{L}, \sin⁡\beta=\frac{\delta Y}{L}
\end{matrix}
\label{eq:2}
\end{equation}
\vspace{-3mm}
    
With the resultant axial deformation $u_l= L-L_0$, the axial force $N$ along the beam can be derived via: 
\vspace{-1mm}
\begin{equation}
    N=\frac{EAu_l}{L_0}
\label{eq:5}
\end{equation}
\vspace{-3mm}

\noindent where $E$ represents the module of elasticity and $A$ denotes the cross-sectional area of the beam. 

\subsubsection{Rotational deflection modeling}

\ 

Apart from the axial deflection, each beam element also undertakes rotational motions at its two nodes. As shown in Fig. \ref{fig:2}, $\theta_1$ and $\theta_2$ are the rotations of nodes 1 and 2, measured from the initial incline axis of the beam element. Thus, the global displacement of each node can be represented by the vector $\theta_1$ and $\theta_2$ and the local nodal rotations become:
\vspace{-1.5mm}
\begin{equation}
    \begin{matrix}
\theta_{1l}=\theta_1+\beta_0-\beta\\
\theta_{2l}=\theta_2+\beta_0-\beta 
\end{matrix}
\label{eq:7}
\vspace{-2mm}
\end{equation}

Using the standard structural analysis results \cite{ref54}, the node moments of the beam (the moment of inertia is represented by $I$) can be related to the local nodal rotations with
\vspace{-1.3mm}
\begin{equation}
   \left \{ \begin{matrix} M_1\\ M_2 \end{matrix}\right \} =\frac{2EI}{L_0}\begin{bmatrix}
  2& 1\\ 1&2\end{bmatrix}\left \{ \begin{matrix}\theta_{1l}\\ \theta_{2l}\end{matrix} \right \} 
\label{eq:8}
\vspace{-1.3mm}
\end{equation}

As a result, for a beam element, its global displacement $(u,w,\theta)$ can be used for calculating the local displacement $(\theta_{1l},\theta_{2l},u_l )$, and the latter can be further utilized to calculate the applied load $(N,M_1,M_2)$  through Eqns. (\ref{eq:1}-\ref{eq:8}). This lays the foundation of the co-rotational modeling \cite{ref51, ref52}.

\subsubsection{Variationally consistent tangent stiﬀness matrix}

\ 

As illustrated in Fig. \ref{fig:3}, we have the following equation as:
\begin{equation}
    \delta u_l=\textbf{e}_1^T \delta \textbf{d}_{21}=\textbf{r} ^T\delta \textbf{p} 
\label{eq:9}
\end{equation}
\noindent with $\delta \textbf{p}$  the variation of the global displacement vector $\textbf{p}=
\begin{small}
    \begin{bmatrix}
u_1&w_1&\theta_1&u_2&w_2&\theta_2 \end{bmatrix}\end{small}^T$ and $\textbf{r}= $ $\begin{bmatrix}
-\cos⁡\beta & -\sin⁡\beta & 0 & \cos\beta & \sin⁡\beta & 0 \end{bmatrix}^T$. When $\delta\alpha$ is small, we have:
\vspace{-1.5mm}
\begin{equation}
\vspace{-1.5mm}
    \delta \alpha =\sin \delta \alpha =\frac{1}{L}\textbf{e}_2^T \delta \textbf{d}_{21}=\frac{1}{L} \textbf{z}^T\delta \textbf{p}                    \label{eq:10}
\vspace{-1.5mm}
\end{equation}
\noindent where $\textbf{z}=
\begin{small}
    \begin{bmatrix}\sin\beta&-\cos\beta& 0& -\sin⁡\beta&\cos⁡\beta&0\end{bmatrix}\end{small}^T$.The changes of rotations can be provided as follows,
\begin{small}
\begin{equation}
    \delta\theta_l = \begin{Bmatrix}
\delta\theta_{1l} \\ \delta\theta_{2l}
\end{Bmatrix}=\begin{Bmatrix}
\delta\theta_{1}-\delta\alpha \\ \delta\theta_{2}-\delta\alpha
\end{Bmatrix}=\left [ T-\frac{1}{L}\left [ \begin{matrix}
 \textbf{z}^T\\\textbf{z}^T\end{matrix} \right ]   \right ] \delta \textbf{p}
 \label{eq:11}
\end{equation}
\end{small}
\noindent where $\begin{footnotesize}T=\begin{bmatrix}
  0& 0 & 1 &  0& 0 &0 \\
 0 &  0& 0 & 0 & 0 &1
\end{bmatrix}\end{footnotesize}$. The change $\delta \textbf{p}_l$ of the local displacement $\textbf{p}_l$ can be described concerning the change $\delta \textbf{p}$ of the global displacement vector $\textbf{p}$ as:
\begin{equation}
        \delta \textbf{p} _l=\left \{ \begin{matrix} \delta u_l \\\delta\theta_{1l} \\\delta\theta _{2l} \end{matrix} \right \}=\textbf{B}\delta \textbf{p}   
\label{eq:12}
\end{equation}
\noindent where \textbf{B} is the transformation matrix in the form of:
\vspace{-2mm}
\begin{equation}
\mathbf{B} =\begin{bmatrix}
 -\cos\beta  & -\sin\beta & 0 & \cos\beta & \sin\beta & 0\\
 -\frac{\sin\beta}{L}  &  \frac{\cos\beta}{L} & 1 & \frac{\sin\beta}{L} & -\frac{\cos\beta}{L} & 0\\
 -\frac{\sin\beta}{L} & \frac{\cos\beta}{L} & 0 & \frac{\sin\beta}{L} & -\frac{\cos\beta}{L} & 1
\end{bmatrix}
\label{eq:13}
\end{equation}
In the local coordinate frame $\{X_l,Y_l\}$, the local internal force vector of the element i can be described by $\textbf{q}_{li}=\begin{footnotesize}
\begin{bmatrix}
 N & M_1 & M_2 
\end{bmatrix}\end{footnotesize}^T$ and the local virtual displacement is $\delta \textbf{p}_{lv}=\begin{footnotesize}\begin{bmatrix}
\delta u_{lv} & \delta \theta_{1lv}& \delta \theta_{2lv}
\end{bmatrix}\end{footnotesize}^T$. In terms of the global coordinate frame $\{X,Y\}$, $\textbf{q}_i$ is the vector of global internal forces for the element $i$ and $\delta \textbf{p}_v$ are the arbitrary virtual displacements. For each beam element, according to the equivalence of virtual work in the local and global systems, we have 
\begin{equation}
    \delta \textbf{p}_v^T\textbf{q}_i=\delta \textbf{p}_{lv}^T \textbf{q}_{li}=(\textbf{B}\delta \textbf{p}_v )^T \textbf{q}_{li}=\delta \textbf{p}_v^T \textbf{B}^T \textbf{q}_{li}
\label{eq:14}
\end{equation}
\noindent Namely,
\vspace{-3mm}
\begin{equation}
\textbf{q}_i = \textbf{B}^T\textbf{q}_{li}
\label{eq:15}
\end{equation}
\noindent Then, the vector $F_{int}$ of internal global forces is provided as
\begin{equation}
\resizebox{.85\hsize}{!}{$\begin{matrix}
\begin{aligned}
&F_{int} = {A_s}_{i=1}^{n_m}q_i =  \\
&{\textstyle \sum_{i=1}^{n_m}} 
\begin{bmatrix}
\cdots & \textbf{q}_{i,1} & \textbf{q}_{i,2} & \textbf{q}_{i,3} & \cdots & \textbf{q}_{i,4} & \textbf{q}_{i,5} & \textbf{q}_{i,6} & \cdots
\end{bmatrix}^T
\end{aligned}
\end{matrix}$}
\label{eq:16}
\end{equation}
where $A_s$ is the assembly index (see Hughes\cite{ref55}). Taking the derivative of $\textbf{q}_i=\textbf{B}^T \textbf{q}_{li}$, we have 
\vspace{-1mm}
\begin{equation}
    \delta \textbf{q}_i=\textbf{k}_i\delta \textbf{p}
\label{eq:17}
\vspace{-1mm}
\end{equation}
in which $\textbf{k}_i$ is the tangent stiffness matrix. The detailed calculations are omitted here and the readers can refer to \cite{ref51, ref52}. Here the final formula of $\textbf{k}_i$ is given as
\begin{equation}
    \textbf{k}_i  =\textbf{B}^T \textbf{C}_l\textbf{B}+\frac{F_N}{L}/\textbf{zz}^T+\frac{M_1+M_2}{L^2}(\textbf{r}\textbf{z}^T+\textbf{zr}^T )  
\label{eq:18}
\end{equation}
\noindent where
\vspace{-4.5mm}
\begin{equation}
\vspace{-3mm}
    \textbf{C}_l= \frac{EA}{L_0} \begin{bmatrix}
 1 &0  &0 \\
 0 & \frac{4I}{A} & \frac{2I}{A}\\
 0 & \frac{2I}{A} & \frac{4I}{A}
\end{bmatrix}
\label{eq:19}
\end{equation}
\noindent Then we can further calculate the global tangent stiffness matrix K of the whole structure as
\begin{equation}
    \textbf{\textit{K}} = {A_s}_{i=1}^{n_m}\textbf{\textit{k}}_i=A_{i=1}^{n_m}\begin{bmatrix}
  & \vdots &  & \vdots & \\
 \cdots & k^i_{1,1} &  \cdots & k^i_{1,2} &  \cdots \\
  & \vdots &  & \vdots & \\
\cdots  &  k^i_{2,1} &  \cdots &  k^i_{2,2} &  \cdots\\
  & \vdots &  & \vdots &
\end{bmatrix}
\label{eq:20}
\end{equation}
\noindent $A_s$ is the assembly operator and $n_m$ is the number of elements. The assembly rows and columns of $\textbf{\textit{k}}_i$ depend on the order of the first and second nodes in the element $i$. Allowing for the node supports, $\textbf{\textit{K}}$ is changed to the modiﬁed global tangent stiﬀness matrix $\mathbf{\textit{\textbf{K}}}_s$. \textcolor{Blue}{\textit{Figure \ref{fig:4} presents typical types of support nodes that are constrained depending on the support conditions. The displacements in the constrained directions are always zero for a particular support node regardless of external forces. The rows and columns of $\textbf{\textit{K}}_s$ concerning this displacement will be zero since the displacement on the support node is still zero. For example, if node $i$ is defined as one support node and fully constrained, then the elements in the related rows $(3i-2,3i-1,3i)$ and related columns $(3i-2,3i-1,3i)$ of $\textbf{\textit{K}}_s$ are all set to be zero.} }
\vspace{-2mm}
\begin{algorithm}[H]
\caption{\textit{\textbf{Member data update}}}
\begin{algorithmic}
\STATE \textbf{\textit{Input:}}
\STATE \noindent $n_{nodes},n_{mem}, m_{conn},\textit{\textbf{A, E, I}}, \textit{\textbf{x}}_0, \textit{\textbf{y}}_0, \textit{\textbf{L}}_0, \boldsymbol{\beta}_0, \textit{\textbf{u}},R_m $
\STATE \textbf{\textit{Calculations:}}
\STATE \textbf{\textit{For}} $i$=1: $n_{mem}$
\STATE \quad\textit{Obtain $\textbf{L}, \cos\beta$, $\cos\beta$ according to Eqns. (1-3)}
\STATE \quad\textit{Obtain $\textbf{q}_{li}$ according to Eqns. (4-8) and thus $\textbf{q}_l$}
\STATE \quad\textit{Obtain $\textbf{\textit{F}}_{int}$ according to Eqns. (13-16)}
\STATE \textbf{\textit{End}}     
\STATE \textit{\textbf{Output:} $\textbf{L}, \textbf{c}, \textbf{s}, \textbf{q}_l, \textbf{F}_{int}$}
\end{algorithmic}
\label{alg:1}
\end{algorithm}
\vspace{-4mm}
\begin{algorithm}[H]
\caption{\textbf{\textit{Tangent stiﬀness matrix update}}}
\begin{algorithmic}
\STATE \textbf{\textit{Input:}}
\STATE \begin{normalsize} $n_{nodes}, n_{mem}, m_{conn}, \textit{\textbf{sup, A, E, I, L, c, s}}, \textit{\textbf{q}}_l, R_m $ \end{normalsize}
\STATE \textbf{\textit{Calculations:}}
\STATE \textbf{\textit{For}} $i$=1: $n_{mem}$
\STATE \ \ \textit{Obtain $\textbf{K}$ according to Eqns. (9-20) }
\STATE \ \ \textit{Obtain \!the \!modified \!$\textbf{\small{K}}_s$\! based\! on the\! support\! condition \!$\textbf{\small{sup}}$}
\STATE \textbf{\textit{End}}     
\STATE \textit{\textbf{Output:} $\textbf{K}_s$}
\end{algorithmic}
\label{alg:2}
\end{algorithm}
\vspace{-6mm}

\subsection{Force-Displacement Modeling}
\begin{figure*}[t]
\vspace{-5mm}
\centerline{\includegraphics[width=1.8\columnwidth]{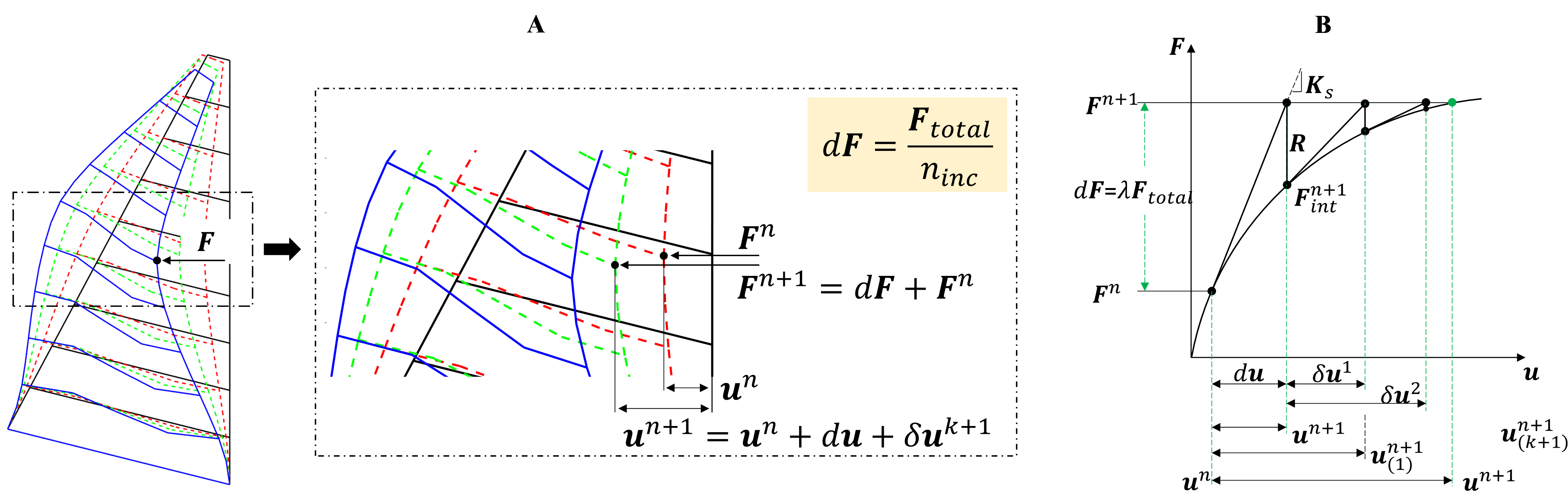}}
\vspace{-3mm}
\caption{One incremental step in the force control. The sketch of finger deformation (A) and the of force-deformation relationship (B).}
\label{Fig:5}
\vspace{-5mm}
\end{figure*}
In this subsection, we describe how to calculate the displacement on a finger of a gripper under external forces. This is an implicit formulation that uses the algorithm of Newton-Raphson iterations at the global level to achieve equilibrium during each incremental load step \cite{ref56}. Different from the linear analysis, the total increment is subdivided into a number of steps, each represented by a cycle in which an equilibrium is reached within a certain tolerance. Specifically, the total vector of externally applied global nodal forces $\textbf{\textit{F}}_{total}$ is defined as a $3n_m\times1$ vector, where the non-zero elements represent the externally applied forces at the selected nodes, allowing for flexible force representation of either concentrated load or distributed loads with no constraints. Here we consider that a total number $n_{inc}$ of load increment steps is required to reach the final equilibrium from the initial equilibrium, and a load of each increment is $d\textbf{\textit{F}}$. Thus, we have
\begin{equation}
\vspace{-1mm}
d\textbf{\textit{F}}=\lambda\textbf{\textit{F}}_{total}
\label{eq:21}
\end{equation}
\noindent with load ratio $\lambda=1/n_{inc}$. Especially, in the $n^{th}$ increment, \vspace{0.4mm}we define the vector of global nodal displacements as $u^n$ and the vector of global nodal forces as $\textbf{\textit{F}}^n$. According to \textbf{Algorithm \ref{alg:1}} and \textbf{Algorithm \ref{alg:2}}, we can calculate $\textit{\textbf{L,c,s}},\textbf{\textit{q}}_{l},\textbf{\textit{F}}_{int}$ and obtain the modified stiffness matrix $\textbf{\textit{K}}_s$. Then, the vector $\textit{d\textbf{u}}$ of each incremental global nodal displacement can be calculated using $\textbf{\textit{K}}_s$ \cite{ref51} as
\vspace{-1.5mm}
\begin{equation}
\vspace{-1.5mm}
d\textbf{\textit{u}}=\textbf{\textit{K}}_s^{-1}d\textbf{\textit{F}}
\label{eq:22}
\end{equation}
As shown in Fig.\ref{Fig:5},  $\textbf{\textit{u}}^n$ and $\textbf{\textit{F}}^n$ can be further updated with
\vspace{-1.5mm}
\begin{equation}
\vspace{-1.5mm}
    \begin{matrix}
\textbf{\textit{u}}^{n+1} = \textbf{\textit{u}}^{n} + d\textbf{\textit{u}}+\delta \textbf{\textit{u}}^{k+1}
 \\
\textbf{\textit{F}}^{n+1} = \textbf{\textit{F}}^n +d\textbf{\textit{F}}
\end{matrix}
\label{eq:23}
\end{equation}
which will later be used in the iteration cycle to achieve equilibrium. Further, we can update $\textbf{\textit{L,c,s}},\textbf{\textit{q}}_l^{n+1},\textbf{\textit{F}}_{int}^{n+1}$ according to \textbf{Algorithm \ref{alg:1}} based on $\textbf{\textit{u}}^{n+1}$. For the iterations, we need to determine the deviation that can be accepted to compare with a set tolerance by defining the residual \textbf{\textit{R}} as 
\begin{equation}
\vspace{-2mm}
    \begin{matrix}
\begin{aligned}
&\textbf{\textit{R}} =  \textbf{\textit{F}}_{int}^{n=1}-\textbf{\textit{F}}^{n+1}
 \\
&R=\sqrt[]{\textbf{\textit{R}}\cdot \textbf{\textit{R}} } 
\end{aligned}
\end{matrix}
\label{eq:24}
\end{equation}
Following the initial preparation process, we enter the iteration cycle, aiming to reach the force equilibrium. A few iteration variables are defined as the iteration number $k=0,tolerance=10^{-3}$. The maximum iteration step is limited by $maxiter=100$. The correction to incremental global nodal displacements is defined as $\delta \textbf{\textit{u}}^k=0$. The temporary vector of local forces in the $k$-th iteration cycle is set as
\vspace{-1.5mm}
\begin{equation}
\vspace{-1mm}
    \textbf{\textit{q}}_{l-temp}^k = \textbf{\textit{q}}_l^{n+1}
\label{eq:25}
\end{equation}
\vspace{-1mm}
\noindent In each step $k$, firstly, the stiffness matrix $\textbf{\textit{K}}_s$ is updated according to \textbf{Algorithm \ref{alg:2}} based on updated current values of inputs and $\textbf{\textit{q}}_{l-temp}^k$. Then the global nodal displacements and member data are updated as
\begin{equation}
\begin{matrix}
\begin{aligned}
&\delta \textbf{\textit{u}}^{k+1} = \delta \textbf{\textit{u}}^k-\textbf{\textit{K}}_s^{-1}\textbf{\textit{R}}
 \\
&\textbf{\textit{u}}_{(k+1)}^{n+1} = \textbf{\textit{u}}^{n} + d\textbf{\textit{u}}+\delta \textbf{\textit{u}}^{k+1}
\end{aligned}
\end{matrix}
\label{eq:26}
\end{equation}
\vspace{-1mm}
\noindent where $\textbf{\textit{u}}_{(k+1)}^{n+1}$ represents $\textbf{\textit{u}}^{n+1}$ in the $(k+1)$-th iteration. When the loop is stopped, $\textbf{\textit{u}}_{(k+1)}^{n+1}$ becomes $\textbf{\textit{u}}^{n+1}$. 

Updating the $\textbf{\textit{q}}_{l-temp}^k,\textbf{\textit{F}}_{int}^{n+1}$ according to \textbf{Algorithm \ref{alg:1}} based on $\textbf{\textit{u}}_{(k+1)}^{n+1}$ is obtained in Eqn. (\ref{eq:25}) and the new residual $\textbf{\textit{R}}$ is updated using Eqn. (\ref{eq:24}). Updating iteration number $k=k+1$, the iteration cycle will terminate if $R\le tolerance$ or $k\ge maxiter$ (in this case, the convergence criteria are not met), and the variables will update to their final value in this $n$-th increment.
\vspace{-1.5mm}
\begin{equation}
\begin{aligned}
\begin{matrix}
\textbf{\textit{q}}^{n+1}_l = \textbf{\textit{q}}^{k+1}_{l-temp}
 \\
\textbf{\textit{u}}^{n+1} = \textbf{\textit{u}}^{n+1}_{(k+1)} = \textbf{\textit{u}}^{n} + d\textbf{\textit{u}}+\delta \textbf{\textit{u}}^{k+1}
\end{matrix}
\end{aligned}
\label{eq:28}
\end{equation}
The complete force-displacement relationship in the $n^{th}$ increment is also illustrated in Figure. \ref{Fig:5}, where the variables updated in both the preliminary step and iteration cycle are demonstrated. The detailed algorithm is illustrated in \textbf{Algorithm. \ref{alg:3}}
\vspace{-1.5mm}
\begin{algorithm}[h]
\caption{\textit{\textbf{Force-displacement Relationship}}}
\begin{algorithmic}
\STATE \textbf{\textit{Input:}}
\STATE $\!\!n_{nodes},n_{mem}, m_{conn},\textit{\textbf{A,E,I}}, \textit{\textbf{x}}_0, \textit{\textbf{y}}_0, \textit{\textbf{L}}_0, \!\boldsymbol{\beta}_0, \textit{\textbf{u}},R_m,\textbf{\textit{F}}_{total}$
\STATE \textbf{\textit{Calculations:}}
\STATE \textbf{\textit{For}} $n$=1: $n_{mem}$

\STATE \quad\textit{calculate $d\textbf{F}$ by Eqn.(\ref{eq:21})}

\STATE \quad\textit{calculate $\textbf{L}, \textbf{c}, \textbf{s}, \textbf{q}_l, \textbf{F}_{int}$ by \small{\textbf{Algorithm \ref{alg:1}}}}

\STATE \quad\textit{calculate $\textbf{K}_s$ by \small{\textbf{Algorithm \ref{alg:2}}}}

\STATE \quad\textit{Obtain $\textbf{\textit{d}}_{u}$ according to Eqn.(\ref{eq:22}) }

\STATE \quad\textit{update $\textbf{u}^{n+1}$ and $\textbf{F}^{n+1}$ by Eqn.(\ref{eq:23})}

\STATE \quad\textit{update $\textbf{L}, \textbf{c}, \textbf{s}, \textbf{q}_l^{n+1}, \!\textbf{F}_{int}^{n+1}$ \!by \small{\textbf{Algorithm \ref{alg:1}}} \normalsize{based on $\textbf{u}^{n+1}$}}

\STATE \quad\textit{calculate the residual $\textbf{R}$ by Eqn.(\ref{eq:24})}

\STATE \quad\textit{set up iteration variables $k, tolerance, maxiter, \delta \textbf{u}$ and $ \textbf{q}_{l-temp}^k$  }

\STATE \quad\textit{
start iterations while $R\! \ge\! tolerance$ and $k\! \le \!maxiter$}

\STATE \qquad\enspace\textit{\footnotesize{\expandafter{\romannumeral1}. 
calculate $\textbf{K}_s$ by \textbf{Algorithm \ref{alg:2}} and $\textbf{q}_{l-temp}^k$ by Eqn.(\ref{eq:25})}}

\STATE \qquad\enspace\textit{\footnotesize{\expandafter{\romannumeral2}. 
update member data $\delta \textbf{u}^{k+1}$ and $\textbf{u}_{(k+1)}^{n+1}$ by Eqn.(\ref{eq:26})
}}

\STATE \qquad\enspace\textit{\footnotesize{\expandafter{\romannumeral3}. 
update $\textbf{q}_{l-temp}^{k+1},\textbf{F}_{int}^{n+1}$ by \textbf{Algorithm \ref{alg:1}} based on $\textbf{u}_{current}$
}}

\STATE \qquad\enspace\textit{\footnotesize{\expandafter{\romannumeral4}. 
calculate the residual $\textbf{R}$ and $R$ by Eqn.(\ref{eq:24})
}}

\STATE \qquad\enspace\textit{\footnotesize{\expandafter{\romannumeral5}. 
update iteration number k=k+1
}}

\STATE \quad\textit{End of while loop iterations}

\STATE \quad\textit{Update variables $\textbf{q}_l^{n+1}$ and $\textbf{u}^{n+1}$ by Eqn. (\ref{eq:28})}

\STATE \textbf{\textit{End}}     
\STATE \textit{\textbf{Output:} $\textbf{q}_l^{n+1},\textbf{u}^{n+1}$}
\end{algorithmic}
\label{alg:3}
\vspace{-1mm}
\end{algorithm}
\vspace{-2mm}

\section{SIMULATION EXPERIMENTS}
{\textcolor{Purple}{Not only to evaluate the performance of the proposed co-rotational method, but also provide insight into Fin-Ray gripper design considering more key design parameters, simulation experiments are conducted and analyzed by comparing the proposed approach with Finite Element Analysis (FEA) in Ansys. FEA is recognized as a crucial benchmark solution for the numerical analysis of mechanical models. Especially for Fin-Ray grippers, research have proven its high accuracy compared with physical experiments, with an average error around 3\% \cite{ref4, ref6, Force_FR}. }

A family of Fin-Ray grippers with various design parameters are considered respectively, and their performances are evaluated based on whether they accurately capture displacements under eternal forces or not. Here a given compliant finger (with width $m$ and Height $n$) is meshed by nodes, and the material parameters are summarized in Table \ref{Table: model parameters}. 
\vspace{-3mm}
\begin{table}[hb]

\caption{\small{\textcolor{Purple}{The model parameters for FEA and the co-rotational method}}}
\vspace{-1mm}
\label{Table: model parameters}
\begin{center}
\vspace{-1mm}
\scalebox{1.1}{\begin{tabular}{lc}
\hline
Item                                      & Value              \\ \hline
Width $m$ (m)                             & 40$e^{-3}$         \\
Height $n$ (m)                            & 72$e^{-3}$         \\
Cross-section of each member ($b, h$) (m) & 20$e^{-3},1e^{-3}$ \\
Young's modulus E (Pa)                      & 2$e^{7}$           \\ \hline
\end{tabular}}
\end{center}
\vspace{-3mm}
\end{table}
\vspace{-1mm}

Four main design parameters are taken into consideration, namely (1) the number of crossbeams jointing the front contact and back contact sides, regardless of \textbf{rigid} or \textbf{soft} crossbeams; (2) the top angle between the front Fin-Ray and back one; (3) the inclination angle of the crossbeams; (4) connection type between crossbeams and front-back Fin-Rays. Specially, we distinguish between two connection types: the crossbeam element freely rotating around the connection is defined as “simple” connection, while the “rigid” connection indicates that crossbeams are fixed with front/back Fin-Rays. 

\begin{figure}[t]
\centerline{\includegraphics[width=0.78\columnwidth]{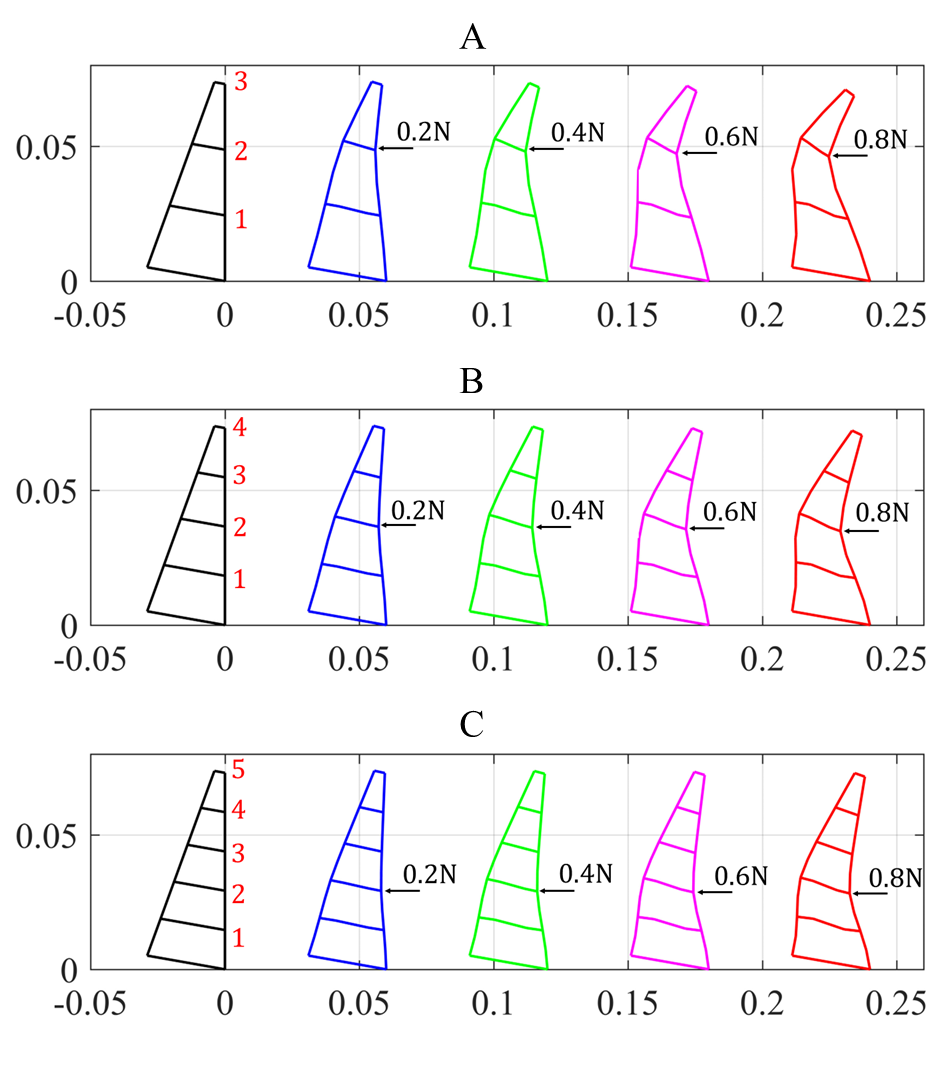}}
\vspace{-5mm}
\caption{\small{The deformations of three Fin-Ray fingers with different numbers of crossbeams such as 2(A), 3(B) and 4(C).}}
\vspace{-2.5mm}
\label{Fig:numbers of crossbeams}
\end{figure}
\begin{table}[t]
\caption{\small{Evaluations on the numbers of crossbeams.}}
\vspace{-3mm}
\label{Table:numbers of crossbeams}
\begin{center}
\begin{threeparttable}[b]
\setlength{\tabcolsep}{2.5mm}{
\scalebox{0.85}{
\begin{tabular}{c|ccccccc}\hline
 & Force & 0.2N & 0.4N & 0.6N & 0.8N &  &  \\ \hline
 & Node & \multicolumn{4}{c}{Error ratio(\%)} & Ave$.^1$(\%) & \multicolumn{1}{l}{SD$.^2$(\%)} \\
 & $1^{st}$ & -9.0 & -11.5 & -12.0 & -13.2 & \cellcolor[HTML]{F9D7AE}11.4 & \cellcolor[HTML]{F9D7AE}1.5 \\
 & $2^{nd}$ & -3.0 & -3.3 & -4.1 & -5.5 & \cellcolor[HTML]{F9D7AE}4.0 & \cellcolor[HTML]{F9D7AE}1.0 \\
 & $3^{rd}$ & 3.9 & 3.6 & 7.3 & N.A$.^3$ & \cellcolor[HTML]{F9D7AE}5.0 & \cellcolor[HTML]{F9D7AE}1.7 \\\multirow{-5}{*}{2} & Ave$.^1$(\%) & \cellcolor[HTML]{ECF4FF}5.3 & \cellcolor[HTML]{ECF4FF}6.1 & \cellcolor[HTML]{ECF4FF}7.8 & \cellcolor[HTML]{ECF4FF}9.35 & \cellcolor[HTML]{D9F5B1}6.8 & \cellcolor[HTML]{D9F5B1}\textbackslash{} \\ \hline
 & Force & 0.2N & 0.4N & 0.6N & 0.8N &  &  \\ \hline
 & Node & \multicolumn{4}{c}{Error ratio(\%)} & Ave$.^1$(\%) & \multicolumn{1}{l}{SD$.^2$(\%)} \\
 & $1^{st}$ & -12.9 & -11.6 & -11.4 & -12.4 & \cellcolor[HTML]{F9D7AE}12.1 & \cellcolor[HTML]{F9D7AE}0.6 \\
 & $2^{nd}$ & -2.6 & -4.0 & -5.3 & -5.8 & \cellcolor[HTML]{F9D7AE}4.4 & \cellcolor[HTML]{F9D7AE}1.2 \\
 & $3^{rd}$ & 4.0 & 1.0 & 1.7 & 0.5 & \cellcolor[HTML]{F9D7AE}1.8 & \cellcolor[HTML]{F9D7AE}1.3 \\
 & $4^{th}$ & 9.9 & 7.5 & 8.3 & 10.2 & \cellcolor[HTML]{F9D7AE}9.0 & \cellcolor[HTML]{F9D7AE}1.1 \\\multirow{-6}{*}{3} & Ave$.^1$(\%) & \cellcolor[HTML]{ECF4FF}7.4 & \cellcolor[HTML]{ECF4FF}6.0 & \cellcolor[HTML]{ECF4FF}6.7 & \cellcolor[HTML]{ECF4FF}7.2 & \cellcolor[HTML]{D9F5B1}6.8 & \cellcolor[HTML]{D9F5B1}\textbackslash{} \\ \hline
 & Force & 0.2N & 0.4N & 0.6N & 0.8N &  &  \\ \hline
 & Node & \multicolumn{4}{c}{Error ratio(\%)} & Ave$.^1$(\%) & \multicolumn{1}{l}{SD$.^2$(\%)} \\
 & $1^{st}$ & -11.4 & -7.9 & -8.5 & -9.2 & \cellcolor[HTML]{F9D7AE}9.3 & \cellcolor[HTML]{F9D7AE}1.3 \\
 & $2^{nd}$ & -2.7 & -1.9 & -2.3 & -4.3 & \cellcolor[HTML]{F9D7AE}2.8 & \cellcolor[HTML]{F9D7AE}0.9 \\
 & $3^{rd}$ & 0.0 & 0.7 & 0.2 & -0.6 & \cellcolor[HTML]{F9D7AE}0.4 & \cellcolor[HTML]{F9D7AE}0.5 \\
 & $4^{th}$ & 7.5 & 5.4 & 6.7 & 6.7 & \cellcolor[HTML]{F9D7AE}6.6 & \cellcolor[HTML]{F9D7AE}0.8 \\
 & $5^{th}$ & 12.0 & 10.1 & 11.1 & 12.1 & \cellcolor[HTML]{F9D7AE}11.0 & \cellcolor[HTML]{F9D7AE}0.8 \\\multirow{-7}{*}{4} & Ave$.^1$(\%) & \cellcolor[HTML]{ECF4FF}6.7 & \cellcolor[HTML]{ECF4FF}5.2 & \cellcolor[HTML]{ECF4FF}5.8 & \cellcolor[HTML]{ECF4FF}6.6 & \cellcolor[HTML]{D9F5B1}6.1 & \cellcolor[HTML]{D9F5B1}\textbackslash{} \\ \hline\end{tabular}
}}
\begin{tablenotes}
     \item \footnotesize{Ave$.^1$ indicate the average value; N.A$.^3$ represents that the FE method \\ is invalid.}
   \end{tablenotes}
\end{threeparttable}
\vspace{-10.5mm}
\end{center}
\end{table}
\vspace{-2mm}
\subsection{The number of crossbeams}
Figure. \ref{Fig:numbers of crossbeams} presents the force-to-displacement simulation of Fin-Ray fingers with 3, 4, or 5 crossbeams. The forces with a magnitude of 0.2 N, 0.4 N, 0.6 N, and 0.8 N are applied at the selected nodes. Regardless of the number of crossbeams, the centering node(s) of the Fin-Ray finger generate(s) the maximum deformation while only slight deformation happens at the peripheral nodes. Considering all three fingers with different numbers of crossbeams, the overall stiffness increases with the rise in the number of crossbeams, providing smaller deformation under same loads. Additionally, we define the force causing collapses of the finger’s structure as the maximum allowable force that prevents the gripper from reaching an unstable status. In this case, the maximum allowable forces of 2-crossbeams), 3-crossbeams, and 4-crossbeams are 0.8 N, 1.2 N, and 1.9 N, respectively.

Further, the discrepancies between the displacements generated from mathematical models and FEA simulations are calculated and listed in Table. \ref{Table:numbers of crossbeams}. 
The model has high accuracy in predicting the deformation at its middle part \textcolor{Blue}{(average error around 3\%, SD around 1\%) }but much larger errors appear near its two ends. Meanwhile, results reveal that the number of crossbeams significantly affects the maximum allowable force that can keep the stability of the structure. A Fin-Ray  gripper with fewer rigid ribs can perform enveloping grasping through wrapping with a heavily deformed Fin-Ray finger. By contrast, the increase in the number of crossbeams enables the gripper to bend smoothly under concentrated loads due to a bigger stiffness, which ensures a stable grasp; Whereas, this can also be a problem when grasping soft or fragile objects as high-stiffness Fin-Ray structure may cause damage. A remarkable finding is that the proposed displacement-force model performs better with the increasing number of crossbeams. Thus, after doing a trade-off of the bent deformation and stiffness of the Fin-Ray structure, we prefer to increase the number of crossbeams to improve the estimation accuracy of displacement and force for the proposed model and avoid the risk of undesirable instabilities. With the influence determined, for simplicity, we choose the gripper with 4 crossbeams in the following experiments.

\vspace{-3mm}
\subsection{The top angle}
Three representative angles are selected (20°, 30°, and 40°). Fig. \ref {Fig:17} reveals the case when force is exerted at node 2, as it generates big enough deformation under a maximum force of 0.8 N. Generally, the maximum deformation under the same force load increases with the increase of top angle, which is true of nodes 1, 2, and 3 except for node 4. In practical application, a large top angle is desirable so that a gripper can deform enough to hold objects. The maximum allowable forces for 40°, 30°, and 20° are 1.8 N, 1.6 N, and 1.2 N, respectively, suggesting a soft gripper with a larger top angle that has better resistance to collapse. However, this will increase the overall weight of the gripper. Thus, optimal design can be conducted in the trade-off of the elasticity, overall size as well as the weight of the gripper. It can be witnessed that the co-rotational theory could well describe Fin-Ray structure with different top angles and achieve relatively accurate displacements. \textcolor{Blue}{Similarly, the error at middle nodes (2 and 3) are smaller (3\%, SD around 1\%) while much larger errors appear near its two ends.}

\vspace{-3mm}
\subsection{The inclination of the crossbeams}

Fin-Ray grippers with various crossbeam inclination angles are considered here. Three representative inclination angles of crossbeams are selected, such as -10°, 0°, and +10°. External force loads with a maximum magnitude of 0.8 N are applied at four nodes in sequence. Similarly, we present the case when forces are exerted at node 2 (see Fig. \ref{Fig:inclination}). For each Fin-Ray finger, the deformation increases with the improvement of force magnitude. Among three fingers, the maximum deformation under the same force load increases when the inclination angle increases, which is true of the cases for nodes 2, 3, and node 1 as an exception.

\begin{figure}[t]
\centerline{\includegraphics[width=0.78\columnwidth]{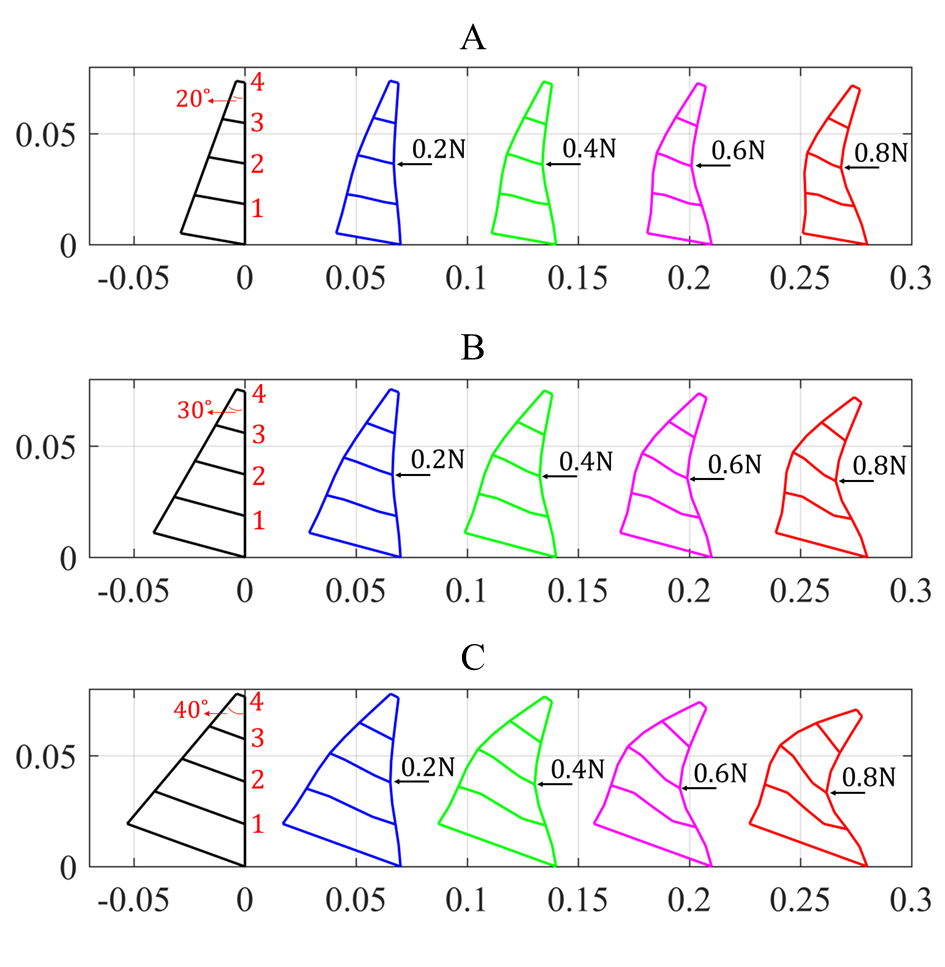}}
\vspace{-5mm}
\caption{\small{The deformations of three Fin-Ray fingers with different top angles such as 20°(A), 30°(B) and 40°(C). The external forces are applied at node 2.}}
\label{Fig:17}
\vspace{-2mm}
\end{figure}
\begin{table}[t]
\caption{\small{Evaluations on the top angle of crossbeams.}}
\vspace{-3mm}
\label{Table: top angles}
\begin{center}
\begin{threeparttable}[b]
\setlength{\tabcolsep}{2.5mm}{
\scalebox{0.85}{\begin{tabular}{c|ccccccc}\hline
 & Force & 0.2N & 0.4N & 0.6N & 0.8N &  &  \\ \hline
 & Node & \multicolumn{4}{c}{Error ratio(\%)} & Ave$.^1$(\%) & \multicolumn{1}{l}{SD$.^2$(\%)} \\
 & $1^{st}$ & -12.9 & -11.6 & -11.4 & -12.4 & \cellcolor[HTML]{F9D7AE}12.1 & \cellcolor[HTML]{F9D7AE}0.6 \\
 & $2^{nd}$ & -2.6 & -4.0 & -5.3 & -5.8 & \cellcolor[HTML]{F9D7AE}4.4 & \cellcolor[HTML]{F9D7AE}1.2 \\
 & $3^{rd}$ & 4.0 & 1.0 & 1.7 & 0.5 & \cellcolor[HTML]{F9D7AE}1.8 & \cellcolor[HTML]{F9D7AE}1.3 \\
 & $4^{th}$ & 9.9 & 7.5 & 8.3 & 10.2 & \cellcolor[HTML]{F9D7AE}9.0 & \cellcolor[HTML]{F9D7AE}1.1 \\\multirow{-6}{*}{20°} & Ave$.^1$(\%) & \cellcolor[HTML]{ECF4FF}7.4 & \cellcolor[HTML]{ECF4FF}6.0 & \cellcolor[HTML]{ECF4FF}6.7 & \cellcolor[HTML]{ECF4FF}7.2 & \cellcolor[HTML]{D9F5B1}6.8 & \cellcolor[HTML]{D9F5B1}\textbackslash{} \\ \hline
 & Force & 0.2N & 0.4N & 0.6N & 0.8N &  &  \\ \hline
 & Node & \multicolumn{4}{c}{Error ratio(\%)} & Ave$.^1$(\%) & \multicolumn{1}{l}{SD$.^2$(\%)} \\
 & $1^{st}$ & -14.1 & -9.3 & -9.5 & -9.5 & \cellcolor[HTML]{F9D7AE}10.6 & \cellcolor[HTML]{F9D7AE}2.0 \\
 & $2^{nd}$ & -3.8 & -4.8 & -5.8 & -6.8 & \cellcolor[HTML]{F9D7AE}5.3 & \cellcolor[HTML]{F9D7AE}1.1 \\
 & $3^{rd}$ & 0.7 & -0.5 & -1.1 & -2.6 & \cellcolor[HTML]{F9D7AE}1.2 & \cellcolor[HTML]{F9D7AE}1.2 \\
 & $4^{th}$ & 4.7 & 3.9 & 5.4 & 8.0 & \cellcolor[HTML]{F9D7AE}5.5 & \cellcolor[HTML]{F9D7AE}1.5 \\\multirow{-6}{*}{30°} & Ave$.^1$(\%) & \cellcolor[HTML]{ECF4FF}5.8 & \cellcolor[HTML]{ECF4FF}4.6 & \cellcolor[HTML]{ECF4FF}5.5 & \cellcolor[HTML]{ECF4FF}6.7 & \cellcolor[HTML]{D9F5B1}5.7 & \cellcolor[HTML]{D9F5B1}\textbackslash{} \\ \hline
 & Force & 0.2N & 0.4N & 0.6N & 0.8N &  &  \\ \hline
 & Node & \multicolumn{4}{c}{Error ratio(\%)} & Ave$.^1$(\%) & \multicolumn{1}{l}{SD$.^2$(\%)} \\
 & $1^{st}$ & -11.7 & -9.4 & -11.5 & -13.3 & \cellcolor[HTML]{F9D7AE}11.5 & \cellcolor[HTML]{F9D7AE}1.4 \\
 & $2^{nd}$ & -5.0 & -6.3 & -7.5 & -9.5 & \cellcolor[HTML]{F9D7AE}7.1 & \cellcolor[HTML]{F9D7AE}1.7 \\
 & $3^{rd}$ & -2.0 & -3.5 & -4.6 & -5.0 & \cellcolor[HTML]{F9D7AE}3.8 & \cellcolor[HTML]{F9D7AE}1.2 \\
 & $4^{th}$ & 2.3 & 0 & 1.0 & 4.4 & \cellcolor[HTML]{F9D7AE}1.9 & \cellcolor[HTML]{F9D7AE}1.6 \\\multirow{-6}{*}{40°} & Ave$.^1$(\%) & \cellcolor[HTML]{ECF4FF}5.3 & \cellcolor[HTML]{ECF4FF}4.8 & \cellcolor[HTML]{ECF4FF}6.2 & \cellcolor[HTML]{ECF4FF}8.1 & \cellcolor[HTML]{D9F5B1}6.1 & \cellcolor[HTML]{D9F5B1}\textbackslash{} \\ \hline\end{tabular}
}
}
\begin{tablenotes}
     \item \footnotesize{Ave$.^1$ indicate the average value.SD$.^2$ represents the standard deviations;}
   \end{tablenotes}
\end{threeparttable}
\vspace{-7mm}
\end{center}
\vspace{-3.5mm}
\end{table}

Besides, the ribs' inclination significantly affects the proposed model's overall behavior. Under the same load at the same node, the Fin-Ray finger with a +10° inclination angle generates the biggest displacement (except for node 1), implying a lower stiffness. The finger with -10° crossbeams has the smallest displacement, revealing a higher stiffness. The instability forces for the three cases are -10° (1.15N), 0°(1.2 N), and +10°(1.35N). Compared to -10°, +10° has a 10\% bigger displacement and 14.5\% higher instability force.

Table. \ref{Table: inclination} presents the discrepancies between the proposed model and FEA, with an overall average error 6\%. The geometrical configurations of crossbeams with different inclinations do not obviously affect the proposed model's accuracy. The gripper with a negative incline angle has a bigger stiffness, resulting in a bigger buckling force. \textcolor{Blue}{For each type, we repeat the loading force at all 4 nodes. The error at nodes 1 and 4 are relatively bigger (around 10\%, SD around 1\%), while middle nodes are smaller (around 3\%, SD around 1\%).}

\vspace{-3.5mm}
\subsection{Connection type} 

The effects of connection types between middle crossbeams and front and back sides of Fin-Rays are simulated. As is demonstrated in Table. \ref{Table: connection types}), the gripper with a simple connection generates almost twice the deformation of the gripper with a rigid connection under the same load. The co-rotational model applied to these two connection types is accurate and resembles very closely. \textcolor{Blue}{The error rate at nodes 1 and 4 are relatively bigger (around 10\%), and nodes 2 and 3 are smaller (around 3\%, SD around 1\%).} The Fin-Ray finger with the simple connection generates almost three-time bigger displacements than one with the rigid connection, undertaking the same load at the same node. In addition, a rigid connection demonstrates a higher maximum allowable force (1.2 N) compared to a simple connection (0.7N). 

\vspace{-2.5mm}
\section{Discussion}
\vspace{-1mm}
\subsection{Design Optimization for Fin-Ray Grippers}
The influence of four parameters is compared and summarized here. First, the inclination of crossbeams is the most decisive parameter that influences the overall performance of both grippers in stiffness and maximum allowable force. A positive inclination angle brings about a more compliant closing (lower stiffness), generating larger deformation under the same load and allowing higher contact force than a negative one. Thus, the priority should be given during the new gripper design. Meanwhile, though simple connections endow the gripper with a lower stiffness, it significantly weakens its strength in resisting instability. In real scenarios, the optimized design should be closely related to the application, carefully considering the trade-off between stiffness and force. 

The proposed modeling accurately captures the finger deformation under external loads, with an average error ratio of 3\% at the middle nodes and overall error 6\%. This is particularly important from an optimization point of view, as the centering nodes will be more used in practical applications. The small average error ratio implies that it is possible to focus the investigation on the finger design with the proposed force-displacement mapping. 

\vspace{-3.5mm}
\subsection{Advantages and Limitations of the Proposed Approach}

The proposed co-rotational method is not a substitute/supplement for existing methods, such as finite elements, pseudo-rigidbody mod \cite{ref4}, discrete-Cosserat-approach \cite{Force_FR}. By contrast, it first constructs the bi-directional relationship between force and displacement and offers an efficient numerical solution for the modeling of compliant grippers within a fair degree of accuracy.
\begin{figure}[t]
\centerline{\includegraphics[width=0.77\columnwidth]{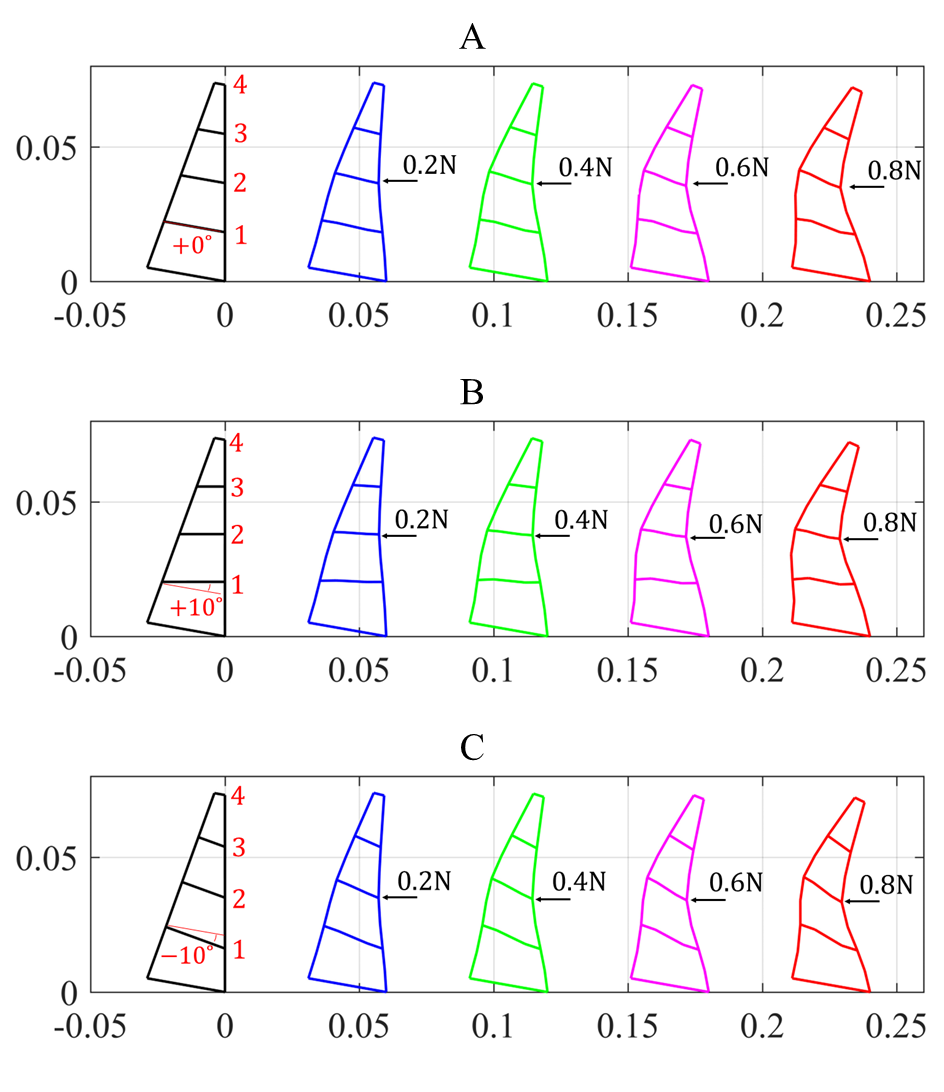}}
\vspace{-4.5mm}
\caption{\small{The deformations of three Fin-Ray fingers with different inclination angles, such as 0°(A), 10°(B) and -10°(C).}}
\label{Fig:inclination}
\vspace{-2.5mm}
\end{figure}
\begin{table}[t]
\caption{Evaluations on the inclination of crossbeams.}
\vspace{-3mm}
\label{Table: inclination}
\begin{center}
\begin{threeparttable}[b]
\setlength{\tabcolsep}{2.5mm}{
\scalebox{0.85}{\begin{tabular}{c|ccccccc}\hline
 & Force & 0.2N & 0.4N & 0.6N & 0.8N &  &  \\ \hline
 & Node & \multicolumn{4}{c}{Error ratio(\%)} & Ave$.^1$(\%) & \multicolumn{1}{l}{SD$.^2$(\%)} \\
 & $1^{st}$ & -12.9 & -11.6 & -11.4 & -12.4 & \cellcolor[HTML]{F9D7AE}12.1 & \cellcolor[HTML]{F9D7AE}0.6 \\
 & $2^{nd}$ & -2.6 & -4.0 & -5.3 & -5.8 & \cellcolor[HTML]{F9D7AE}4.4 & \cellcolor[HTML]{F9D7AE}1.2 \\
 & $3^{rd}$ & 4.0 & 1.0 & 1.7 & 0.5 & \cellcolor[HTML]{F9D7AE}1.8 & \cellcolor[HTML]{F9D7AE}1.3 \\
 & $4^{th}$ & 9.9 & 7.5 & 8.3 & 10.2 & \cellcolor[HTML]{F9D7AE}9.0 & \cellcolor[HTML]{F9D7AE}1.1 \\\multirow{-6}{*}{0°} & Ave$.^1$(\%) & \cellcolor[HTML]{ECF4FF}7.4 & \cellcolor[HTML]{ECF4FF}6.0 & \cellcolor[HTML]{ECF4FF}6.7 & \cellcolor[HTML]{ECF4FF}7.2 & \cellcolor[HTML]{D9F5B1}6.8 & \cellcolor[HTML]{D9F5B1}\textbackslash{} \\ \hline
 & Force & 0.2N & 0.4N & 0.6N & 0.8N &  &  \\ \hline
 & Node & \multicolumn{4}{c}{Error ratio(\%)} & Ave$.^1$(\%) & \multicolumn{1}{l}{SD$.^2$(\%)} \\
 & $1^{st}$ & -8.1 & -8.2 & -9.2 & -10.0 & \cellcolor[HTML]{F9D7AE}8.9 & \cellcolor[HTML]{F9D7AE}0.8 \\
 & $2^{nd}$ & -3.5 & -1.3 & -3.2 & -3.7 & \cellcolor[HTML]{F9D7AE}2.9 & \cellcolor[HTML]{F9D7AE}1.0 \\
 & $3^{rd}$ & 1.2 & 3.4 & 2.6 & 1.9 & \cellcolor[HTML]{F9D7AE}2.3 & \cellcolor[HTML]{F9D7AE}0.8 \\
 & $4^{th}$ & 11.5 & 6.8 & 8.6 & 11.3 & \cellcolor[HTML]{F9D7AE}9.6 & \cellcolor[HTML]{F9D7AE}2.0 \\\multirow{-6}{*}{+10°} & Ave$.^1$(\%) & \cellcolor[HTML]{ECF4FF}6.1 & \cellcolor[HTML]{ECF4FF}4.9 & \cellcolor[HTML]{ECF4FF}5.9 & \cellcolor[HTML]{ECF4FF}6.8 & \cellcolor[HTML]{D9F5B1}5.9 & \cellcolor[HTML]{D9F5B1}\textbackslash{} \\ \hline
 & Force & 0.2N & 0.4N & 0.6N & 0.8N &  &  \\ \hline
 & Node & \multicolumn{4}{c}{Error ratio(\%)} & Ave$.^1$(\%) & \multicolumn{1}{l}{SD$.^2$(\%)} \\
 & $1^{st}$ & -9.2 & -9.7 & -11.4 & -12.6 & \cellcolor[HTML]{F9D7AE}10.7 & \cellcolor[HTML]{F9D7AE}1.4 \\
 & $2^{nd}$ & -1.1 & -3.3 & -4.6 & -5.4 & \cellcolor[HTML]{F9D7AE}3.6 & \cellcolor[HTML]{F9D7AE}1.6 \\
 & $3^{rd}$ & 3.1 & 1.8 & 1.8 & 0.6 & \cellcolor[HTML]{F9D7AE}1.8 & \cellcolor[HTML]{F9D7AE}0.9 \\
 & $4^{th}$ & 9.1 & 5.8 & 8.3 & 8.5 & \cellcolor[HTML]{F9D7AE}7.9 & \cellcolor[HTML]{F9D7AE}1.3 \\\multirow{-6}{*}{-10°} & Ave$.^1$(\%) & \cellcolor[HTML]{ECF4FF}5.6 & \cellcolor[HTML]{ECF4FF}5.2 & \cellcolor[HTML]{ECF4FF}6.5 & \cellcolor[HTML]{ECF4FF}6.8 & \cellcolor[HTML]{D9F5B1}6.1 & \cellcolor[HTML]{D9F5B1}\textbackslash{} \\ \hline\end{tabular}
}
}
\begin{tablenotes}
     \item \footnotesize{Ave$.^1$ indicate the average value. SD$.^2$ represents the standard deviations;}
   \end{tablenotes}
\end{threeparttable}
\end{center}
\vspace{-7.5mm}
\end{table}
One of the main novelties of lies that in the fair accuracy and generality of the co-rotational theory, allowing considering both axial deformation and rotational deformation of single beam element. We achieve fair accuracy 6\% in predicting displacements, compared to 3.7\% in \cite{Force_FR} and 3.07\% in \cite{ref4}. Meanwhile, traditional methods tend to simplify the modeling by assuming the ribs as in-extensible beam and neglect the axial deformation of this elements \cite{ref16}. The proposed algorithm can extended to both rigid and soft robots. Especially, a soft/continuum gripper can be considered a Fin-Ray gripper with numerous crossbeams and low stiffness material for readily deforming. While a rigid gripper may be regarded as a Fin-Ray gripper with high-stiff materials. 

Besides, in some specific cases, the computational times are strongly reduced. To ensure more accurate simulation results, the large deformation option should be enabled in FEA. Our approach simplifies the description of the model with reasonable nodes and meshes. It takes an average number of 3 iterations to converges each load increment, faster than commercial FEA simulation in Ansys.


\vspace{-1mm}
\section{CONCLUSIONS}
This article first constructs a bidirectional model that mutually maps finger deformation and external contact force, facilitating further design optimization (Part I) and force-aware grasping (Part II). In this part, the theoretical modeling of beam element employing the co-rotational concept is presented, with few assumptions needed and high computational efficiency. A force-displacement relationship depicting finger deformation under external forces is further proposed, and experimentally validated in simulations. Results reveal that our model has an overall average error ratio around 6\% compared with FEA. In particular, insights into the influence of four key design parameters for Fin-Ray grippers' performance have been provided, facilitating future works regarding gripper design optimization. Part II of this article presents a displacement-force relationship and influence of modeling parameters, providing further verification of the proposed model and a novel mathematical tool for force control of compliant grippers with no reliance on force sensor.
\begin{figure*}[htbp]
\centerline{\includegraphics[width=1.5\columnwidth]{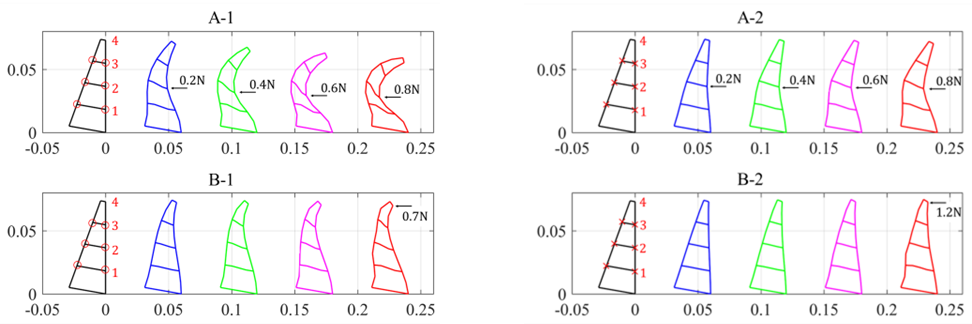}}
\vspace{-1mm}
\caption{\small{The deformations of Fin-Ray fingers with two connection types, including “simple” (A-1, B-1) and “rigid” (A-2, B-2) connections.}}
\label{Fig:18}
\vspace{-2.5mm}
\end{figure*}
\begin{table*}[!thp]
\caption{\small{Evaluations on the two connection types for the Fin-Ray finger.}}
\vspace{-2.5mm}
\label{Table: connection types}
\begin{center}
\begin{threeparttable}[b]
\setlength{\tabcolsep}{2.3mm}{
\scalebox{0.88}{\begin{tabular}{c|ccccccc|c|ccccccc}\hline
 & Force & 0.2N & 0.4N & 0.6N & 0.8N &  &  &  & Force & 0.2N & 0.4N & 0.6N & 0.8N &  &  \\ \hline
 & Node & \multicolumn{4}{c}{Error ratio(\%)} & A$.^2$(\%) & \multicolumn{1}{l|}{SD$.^3$(\%)} &  & Node & \multicolumn{4}{c}{Error ratio(\%)} & A$.^2$(\%) & \multicolumn{1}{l}{SD$.^3$(\%)} \\
 & $1^{st}$ & -5.8 & -9.2 & -10.5 & -11.5 & \cellcolor[HTML]{F9D7AE}9.3 & \cellcolor[HTML]{F9D7AE}2.2 &  & $1^{st}$ & -12.9 & -11.6 & -11.4 & -12.4 & \cellcolor[HTML]{F9D7AE}12.1 & \cellcolor[HTML]{F9D7AE}0.6 \\
 & $2^{nd}$ & -1.9 & -3.5 & -3.5 & -3.2 & \cellcolor[HTML]{F9D7AE}3.0 & \cellcolor[HTML]{F9D7AE}0.7 &  & $2^{nd}$ & -2.6 & -4.0 & -5.3 & -5.8 & \cellcolor[HTML]{F9D7AE}4.4 & \cellcolor[HTML]{F9D7AE}1.2 \\
 & $3^{rd}$ & 0.2 & -0.5 & -0.9 & 0.0 & \cellcolor[HTML]{F9D7AE}0.4 & \cellcolor[HTML]{F9D7AE}0.4 &  & $3^{rd}$ & 4.0 & 1.0 & 1.7 & 0.5 & \cellcolor[HTML]{F9D7AE}1.8 & \cellcolor[HTML]{F9D7AE}1.3 \\
 & \cellcolor[HTML]{9B9B9B}$4^{th}$ & \cellcolor[HTML]{9B9B9B}3.8 & \cellcolor[HTML]{9B9B9B}6.6 & \cellcolor[HTML]{9B9B9B}9.9 & \cellcolor[HTML]{9B9B9B}N.A$.^4$ & \cellcolor[HTML]{9B9B9B}6.8 & \cellcolor[HTML]{9B9B9B}2.5 &  & $4^{th}$ & 9.9 & 7.5 & 8.3 & 10.2 & \cellcolor[HTML]{F9D7AE}9.0 & \cellcolor[HTML]{F9D7AE}1.1 \\\multirow{-6}{*}{S$.^1$} & A$.^2$(\%) & \cellcolor[HTML]{ECF4FF}2.9 & \cellcolor[HTML]{ECF4FF}5.0 & \cellcolor[HTML]{ECF4FF}6.2 & \cellcolor[HTML]{ECF4FF}4.9 & \cellcolor[HTML]{D9F5B1}4.9 & \cellcolor[HTML]{D9F5B1}\textbackslash{} & \multirow{-6}{*}{F$.^1$} & A$.^2$(\%) & \cellcolor[HTML]{ECF4FF}7.4 & \cellcolor[HTML]{ECF4FF}6.0 & \cellcolor[HTML]{ECF4FF}6.7 & \cellcolor[HTML]{ECF4FF}7.2 & \cellcolor[HTML]{D9F5B1}6.8 & \cellcolor[HTML]{D9F5B1}\textbackslash{} \\ \hline\end{tabular}
}
}
\begin{tablenotes}
     \item \footnotesize{S$.^1$ and F$.^1$ represent “Simple” and “Rigid”, respectively. A$.^2$ indicates the average values; SD$.^3$ represents the standard deviations; N.A.$^4$ indicates that the FE method in Ansys fails.}
   \end{tablenotes}
\end{threeparttable}
\end{center}
\vspace{-6.5mm}
\end{table*}
\vspace{-1mm}
\AtNextBibliography{\small}
\printbibliography

\end{document}